\documentclass{article}

\PassOptionsToPackage{numbers, compress}{natbib}

\usepackage[dblblindworkshop,final]{neurips_2026}

\workshoptitle{Linguistic Principles for Foundation Models}

\usepackage{microtype}
\usepackage{hyperref}
\usepackage{url}
\usepackage{booktabs}

\usepackage[utf8]{inputenc} 
\usepackage[T1]{fontenc}    
\usepackage{amsfonts}       
\usepackage{nicefrac}       
\usepackage{xcolor}         
\usepackage{graphicx}

\definecolor{darkblue}{rgb}{0, 0, 0.5}
\hypersetup{colorlinks=true, citecolor=darkblue, linkcolor=darkblue, urlcolor=darkblue}

\usepackage{amssymb}
\usepackage{amsmath}
\usepackage{float}
\usepackage{multirow}
\usepackage{makecell}
\usepackage{bm}
\usepackage{subcaption}
\usepackage{tcolorbox}
\tcbuselibrary{breakable, skins}
\usepackage{stfloats}
\usepackage{pifont}

\definecolor{darkgreen}{rgb}{0, 0.45, 0}
\definecolor{darkred}{rgb}{0.65, 0, 0}
\newcommand{\cmark}{\textcolor{darkgreen}{\ding{51}}}
\newcommand{\xmark}{\textcolor{darkred}{\ding{55}}}

\title{Test-Time Augmentation for LLMs:\\ Where to Inject Diversity at Inference Time}

\author{%
  Nikita Kozodoi \\
  Amazon Web Services \\
  \texttt{kozodoi@amazon.com} \\
  \And
  Zainab Afolabi \\
  Amazon Web Services \\
  \texttt{zafolabi@amazon.com} \\
  \And
  Jack Butler \\
  Amazon Web Services \\
  \texttt{jackbtlr@amazon.com} \\
}

\begin{document}

\maketitle

\begin{abstract}
Aggregating several LLM responses to the same question can improve accuracy, but every method that does so must decide where the responses should differ. Established test-time scaling methods such as self-consistency vary only the reasoning path, resampling from a fixed input. The input modification offers several further options: the wording of the question can be rephrased, the spelling can be perturbed, and the accompanying images can be transformed. We study Test-Time Augmentation (TTA), which aggregates predictions across transformed versions of the input, and use it to ask where diversity is best injected. Comparing these options across six benchmarks at a matched number of inference calls, we find that where the diversity comes from matters more than how much of it there is. Meaning-preserving paraphrases of the input yield the largest accuracy gains (+1.80 pp over a single call) and reach peak accuracy with the fewest samples. Yet, they are not the most diverse: character-level noise makes the responses disagree far more often (37.8\% versus 27.4\%) while gaining less accuracy, and perturbing two modalities at once falls below the single-call baseline. Not all diversity is equally useful: perturbations that keep the meaning intact produce disagreement that aggregation can use, while those that corrupt it produce disagreement that is wrong. Overall, our results suggest that for mid-tier LLMs such as Claude Haiku, varying how a question is phrased is a better use of an inference call than varying the reasoning path. Our TTA implementation is available at \url{github.com/aws-samples/sample-genai-reflection-for-bedrock}.
\end{abstract}

%
%

%
%

\section{Introduction}
\label{sec_introduction}

Large Language Models (LLMs) demonstrate strong performance across diverse tasks, from mathematical reasoning to multi-modal understanding, yet their answers depend not only on what a question asks but on how it is worded. Semantically equivalent phrasings of the same question can yield different predictions, a sensitivity to surface form documented across punctuation, instruction perturbations, and paraphrase type \citep{seleznyov2025punctuation, chatziveroglou2025reasoning, wahle2024paraphrasetypes}. Much of the literature measures this effect in order to reduce it, treating it as a weakness to be fixed. A second line of work instead puts it to use, rephrasing inputs to improve accuracy on individual tasks \citep{deng2024rar, zhou2024scop, yadav2024pag}. We follow that view: if wording moves predictions, varying it on purpose yields responses whose errors are less correlated than repeated samples of one phrasing, and aggregating turns that spread into accuracy improvements.

\begin{figure*}[t]
\centering
\includegraphics[width=\textwidth]{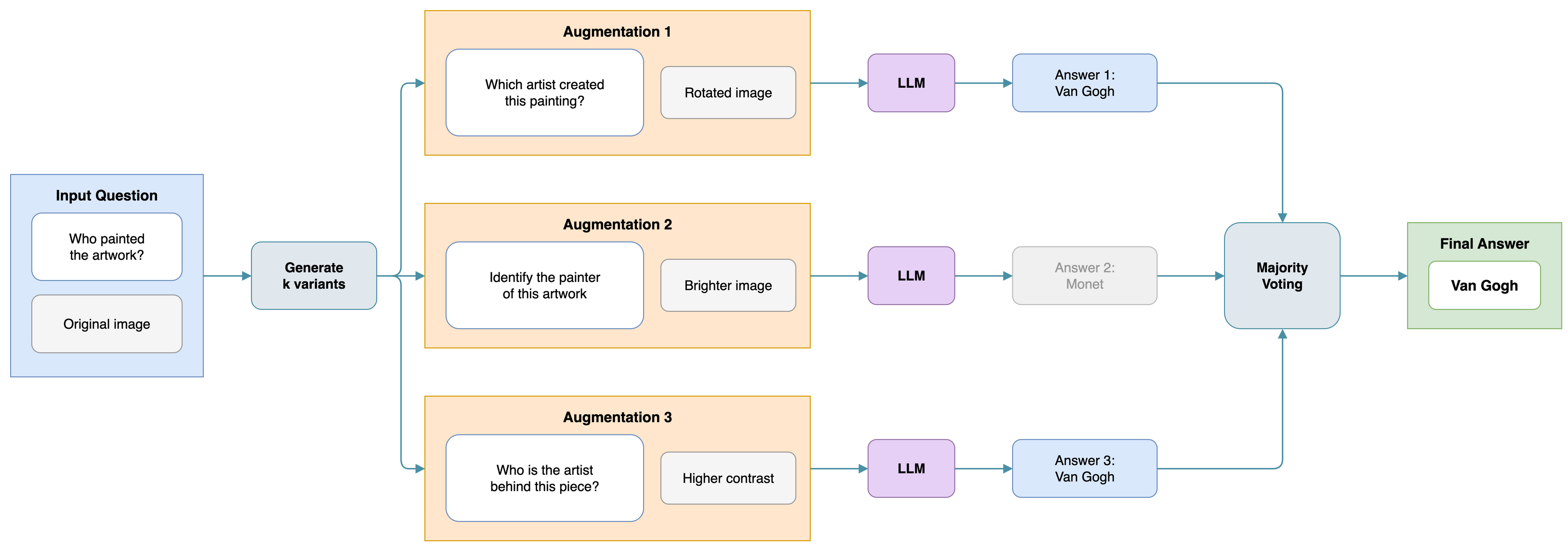}
\caption{TTA framework. The input question and image are augmented into $k$ variants through text paraphrasing and image transformations. Each variant is processed independently by the LLM, producing answer candidates that are aggregated via majority voting.}
\label{fig:tta_framework}
\end{figure*}

This is the intuition behind Test-Time Augmentation (TTA), a well-established technique in supervised learning that aggregates predictions across multiple transformed versions of a test input \citep{shanmugam2020tta}. In computer vision, TTA combines predictions over rotated, flipped, or cropped images, and prior work has shown smaller gains in NLP applications such as text classification \citep{lu2022tta} and factual probing \citep{kamoda2023tta}. Despite its success in supervised settings, TTA has not been systematically studied for generative LLMs. Applied to LLMs, it generates multiple variants of an input, such as paraphrases of a question, and aggregates predictions over them via majority voting, as Figure~\ref{fig:tta_framework} illustrates for multi-modal question answering. TTA is training-free and independent of the choice of base model, making it straightforward to layer onto existing inference pipelines.

Most inference-time scaling techniques sample, refine, or restructure the model's output: examples include self-consistency \citep{wang2022selfconsistency}, which samples multiple reasoning paths from chain-of-thought (CoT) prompting \citep{wei2022cot}, Self-Refine \citep{madaan2023selfrefine}, and Tree of Thoughts \citep{yao2023treethoughtsdeliberateproblem}. These methods spend the entire compute budget on the output side, repeatedly re-running the model on the same input. TTA can be viewed as extending self-consistency: rather than sampling repeatedly from a fixed input, it also perturbs the input before sampling, adding input-side diversity on top of output-side diversity. Both approaches spend the same thing, an extra inference call, and each call is a chance to vary something. The reasoning path is therefore only one of several places diversity can enter. On the input side alone, an augmentation can reword the question while keeping its meaning, disturb its spelling, or transform an accompanying image, and there is no reason to expect the resulting variation to be equally useful in each case. The techniques we study, paraphrasing, character-level noise, and image transforms, are all established, so our comparison focuses on where diversity is injected rather than on any particular augmentation design. Prior work has applied paraphrase aggregation to narrow tasks such as mathematical reasoning \citep{zhou2024scop} and intent classification \citep{yadav2024pag}, but to our knowledge none systematically compares these options against one another at a matched inference budget.

This paper investigates the following question. \emph{Given a fixed number of inference calls, where should diversity be injected: into the meaning of the input, its surface form, its non-text modalities, or the reasoning path?} We answer it on MMLU, MMMLU, MMMU, HLE, Math500, and IMDB Reviews.

Our contributions are three-fold. First, we place several ways of diversifying responses in direct competition under one framework, benchmarking semantic, lexical, visual, and combined augmentation against CoT prompting and self-consistency at a matched number of inference calls. Second, we show that where the diversity comes from matters more than how much of it there is: meaning-preserving paraphrases give the largest gains and reach peak accuracy with the fewest samples, yet character-level noise produces markedly more candidate disagreement while gaining less, and perturbing two modalities at once falls below the single-call baseline. We show that not all diversity is equally useful: the perturbations that keep the meaning intact are the ones that aggregation can benefit from. Third, we ablate the number of augmentations, the augmented modality, the rephrasing model, and the base model, establishing when varying the input is worth its extra inference calls.


\section{Related Work}
\label{sec_related_work}

A growing body of work scales test-time compute to improve LLM accuracy without retraining. \citet{snell2025scaling} show that repeated sampling and verification can outperform scaling model parameters, while \citet{butler2025finding} examine cost-quality-speed trade-offs in iterative reflection. Self-Refine \citep{madaan2023selfrefine} and self-debugging \citep{chen2024teaching} use iterative self-critique, although the reliability of self-verification is contested \citep{stechly2024selfverificationlimitationslargelanguage, valmeekam2023largelanguagemodelsreally}. Structured reasoning approaches such as Tree of Thoughts \citep{yao2023treethoughtsdeliberateproblem}, Graph of Thoughts \citep{Besta_2024}, and process reward models \citep{lightman2024lets, setlur2025rewarding} modify the reasoning process itself, and test-time training \citep{akyürek2025surprisingeffectivenesstesttimetraining, hubotter2025efficiently} adapts model parameters during inference. Unlike these approaches, TTA operates only on input representations and requires neither parameter updates nor self-verification.

Among output-side methods, self-consistency \citep{wang2022selfconsistency} is the most widely used baseline, with extensions including Mirror-Consistency \citep{huang2025mirror} and confidence-weighted voting \citep{taubenfeld2025confidence}. The input-side regime we study is motivated by the well-documented sensitivity of LLMs to prompt phrasing \citep{seleznyov2025punctuation, chatziveroglou2025reasoning, agrawal2025robustness, wahle2024paraphrasetypes}, where minor surface changes shift predictions substantially, and which has been quantified into dedicated indices \citep{chatterjee2024posix}. Closest to the use we make of it, \citet{cox2025mapping} sample across paraphrases to improve uncertainty calibration, treating the spread over semantically equivalent prompts as signal rather than noise, though they target calibration rather than accuracy. One study disagrees with the premise: \citet{hua2025flaw} argue that much reported prompt sensitivity is an artifact of rigid answer matching and disappears under LLM-as-a-judge evaluation. Our setting is not subject to that critique, since we compare methods under one fixed extraction procedure, and the disagreement we measure is between a model's own answers to different phrasings rather than between scoring schemes.

The methods that already exploit input rephrasings each fix one way of perturbing the input. \citet{deng2024rar} propose Rephrase-and-Respond (RaR), which uses a single rephrasing for clarification rather than aggregation, whereas semantic TTA generates $k$ rephrasings and majority-votes over them. SCoP \citep{zhou2024scop} and PAG-LLM \citep{yadav2024pag} do aggregate paraphrases, but on one task each and against task-specific baselines, CAPE \citep{jiang2023cape} ensembles augmented prompts for calibration rather than accuracy, and \citet{kamoda2023tta} apply TTA to factual probing with mixed results. In computer vision, the closest analog is ZERO \citep{farina2024zero}, which augments visual inputs $N$ times for vision-language models. Our contribution is to bring these scattered options into one framework and ask which kind of perturbation makes the best use of each inference call.

A separate line of research uses multiple distinct prompts through prompt ensembling: boosted ensembles \citep{pitis2023boosted}, multi-prompt decoding \citep{guo2024mped}, and PREFER \citep{zhang2023prefer}, all of which typically require prompt optimization or selection. Model ensembles \citep{ai2025beyondmajority, niimi2024sentiment} combine outputs from multiple models, which is often impractical at deployment scale. In contrast, TTA generates rephrasings on-the-fly without task-specific prompt engineering. Augmentation has also been extensively studied at training time \citep{chai2025textda, nllb2022, trung-etal-2024-reft, yao2025mend, choi2025roparq}; our work differs by applying augmentation purely at inference time, with no parameter updates or additional training data.

\section{Test-Time Augmentation for LLMs}
\label{sec_methods}

\subsection{Framework Overview}

Given an input query $x$ and an LLM $f$, standard inference produces a single response $y = f(x)$. TTA instead generates $k$ augmented versions $\{x_1, \ldots, x_k\}$, predicts on each, and aggregates.

For tasks with discrete answers, we aggregate responses by majority voting. Let $\{y_1, \ldots, y_k\}$ denote the predictions for the augmented inputs. The final prediction is:
\begin{equation}
\hat{y} = \arg\max_{c} \sum_{i=1}^{k} \mathbf{1}[y_i = c]
\end{equation}
where $\mathbf{1}[\cdot]$ is the indicator function and $c$ ranges over answer choices. Ties are broken at random.

The augmentation function is what decides where the diversity comes from, and it is the design choice we study. We consider three augmentations, each perturbing a different aspect of the input: semantic TTA changes the wording while keeping the meaning, lexical TTA changes the spelling, and visual TTA changes an accompanying image. Self-consistency serves as the output-side counterpart that leaves the input untouched. Table~\ref{tab:augmentation_examples} shows examples for the text-based strategies.

\begin{table}[!t]
  \centering
  \small
  \caption{Examples of inputs for each method. CoT prompting is the single-call baseline; TTA varies the input (input-side diversity); self-consistency feeds the same input $k$ times and varies only the reasoning path (output-side diversity). Typos in lexical TTA are intentional.}
  \label{tab:augmentation_examples}
  \begin{tabular*}{\textwidth}{@{\extracolsep{\fill}}lllp{0.42\textwidth}@{}}
  \toprule
  Method & Diversity & Transformation & Input example \\
  \midrule
  CoT prompting & None & None & What is the capital of France? \\
  \midrule
  \multirow{2}{*}{Semantic TTA} & \multirow{2}{*}{Input} & \multirow{2}{*}{Paraphrasing} & Which city serves as France's capital? \\
                                & & & Tell me the capital city of France \\
  \midrule
  \multirow{2}{*}{Lexical TTA}  & \multirow{2}{*}{Input} & \multirow{2}{*}{Character noise} & What is the captial of Frnace? \\
           & & & Whatt is the capital of France? \\
  \midrule
  \multirow{2}{*}{Self-consistency} & \multirow{2}{*}{Output} & \multirow{2}{*}{None} & What is the capital of France? \\
           & & & What is the capital of France? \\
  \bottomrule
  \end{tabular*}
\end{table}

\subsection{Semantic TTA}

Semantic TTA generates paraphrased versions of the input that preserve meaning while varying surface form. Given a question $x$, we prompt an LLM to produce $k$ semantically equivalent rephrasings $\{x_1^{\text{sem}}, \ldots, x_k^{\text{sem}}\}$ in a single LLM call. Each rephrasing is then answered independently and predictions are aggregated by majority voting. This exploits the well-documented sensitivity of LLMs to prompt phrasing \citep{seleznyov2025punctuation, chatziveroglou2025reasoning}: aggregating across phrasings reduces variance from any single surface form. The rephrasing prompt instructs the LLM to preserve meaning, intent, and answer format while varying vocabulary and sentence structure (see Appendix~\ref{app:prompts}).

The diversity of the augmented inputs depends on the rephraser model and its sampling temperature. We treat both as design choices, including using a stronger model than the answering model to generate the rephrasings, which we evaluate in Appendix~\ref{app:rephraser}. Producing all $k$ rephrasings in a single call keeps the augmentation overhead small relative to the $k$ answer calls. To ensure the rephrasings remain answerable, we explicitly instruct the rephraser to preserve answer choices, formatting requirements, and references to images so that the question stays well-posed under aggregation.

\subsection{Lexical TTA}

Lexical augmentation applies character-level perturbations to the input, requiring no additional LLM call for augmentation and thus being computationally cheaper. We apply three transformation types: random character swaps within words, character insertions and deletions, and injection of typos and spelling mistakes. We use a perturbation probability of 5\% per word, capped at 10 perturbations per question: prior work on character-level noise \citep{agrawal2025robustness} suggests that higher rates degrade comprehension, while lower rates produce near-identical inputs.

Lexical TTA is motivated by the observation that LLMs are largely robust to small character-level corruption such as typos and reordering, so perturbed inputs should yield correct answers most of the time while still inducing diversity in the model's predictions. The trade-off is that the perturbations preserve surface form rather than meaning: they can shift predictions on borderline questions where the perturbed token coincides with a content word. As a result, lexical TTA is essentially free to apply but is expected to provide weaker gains than semantic TTA, a hypothesis we verify in Section~\ref{sec_main_results}.

\subsection{Visual TTA}

For multi-modal tasks involving images, we extend TTA to visual transformations of the image inputs while keeping the textual question unchanged. We apply three transformation types: small-angle rotation, brightness adjustments, and contrast adjustments. The rotation range $\pm \alpha^\circ$ and the photometric ranges $\pm \beta\%$ act as hyperparameters that control augmentation strength. 

Following common practice in visual TTA \citep{shanmugam2020tta}, we keep transformations mild to preserve the semantic content of the image: small rotations and modest brightness or contrast shifts produce visibly distinct yet recognizable variants of the original. For each test input, we sample $k$ independent transformation parameter triples and apply them to obtain $k$ augmented images, each paired with the original question. The resulting $k$ predictions are aggregated by majority voting. We also explore combining visual and text-based augmentations on multi-modal benchmarks.

\subsection{Self-Consistency}

Self-consistency \citep{wang2022selfconsistency} is one of the most established inference-time scaling methods. Starting from CoT prompting \citep{wei2022cot}, it samples $k$ reasoning paths from the same input at temperature $T>0$ and aggregates the resulting answers by majority voting. Concretely, given input $x$, we sample $k$ responses $\{y_1, \ldots, y_k\}$ via independent inference calls at $T=0.75$ with CoT prompting, then majority-vote over the extracted answers. Since every TTA variant we study samples at the same temperature, its answers carry the same output-side diversity as self-consistency plus the input-side diversity from augmentation, so comparing the two at matched $k$ isolates the contribution of input variation.

%
%

\section{Experimental Setup}
\label{sec_setup}


\subsection{Datasets}

We evaluate TTA across six established LLM benchmarks. We chose datasets that allow us to vary the things an augmentation could act on: the answer format, the language of the question, and whether the input is text alone. \textbf{MMLU} \citep{hendrycks2021measuring} tests multitask language understanding across 57 subjects, and its multilingual extension \textbf{MMMLU} \citep{openai2024mmmlu}, translated into 14 languages, lets us assess TTA robustness beyond English. \textbf{MMMU} \citep{yue2024mmmu} requires joint image and text reasoning, while \textbf{HLE} \citep{phan2025hle} poses expert-crafted questions across math, sciences, and humanities. \textbf{Math500} \citep{lightman2024lets} covers math problems spanning algebra, arithmetic, geometry, and calculus, and \textbf{IMDB Reviews} \citep{maas-EtAl:2011:ACL-HLT2011} is a binary sentiment classification task over movie reviews. 

Together the datasets we use span multiple-choice, free-form mathematical, and binary answers, which lets us identify the methods that work well across answer formats rather than on any single one. We use a randomly sampled subset of 400 examples per dataset, balancing coverage with computational cost. Statistical significance testing is reported in Section~\ref{sec_main_results}.

\subsection{Models and Methods}

We conduct experiments on the Claude 4.5 model family, using Claude 4.5 Haiku \citep{claude45} as the primary model. All methods produce answers via the same CoT-prompting template at $T=0.75$ (see Appendix~\ref{app:prompts:tta}). Apart from the baseline, each diversity injection method aggregates $k$ independent responses by majority voting, and they differ only in how those $k$ candidates are generated. \textbf{CoT prompting} \citep{wei2022cot} is the single-call baseline with no aggregation. \textbf{Self-consistency} \citep{wang2022selfconsistency} feeds the same input to the LLM $k$ times. \textbf{Semantic TTA} generates $k$ rephrasings in one additional LLM call. \textbf{Lexical TTA} creates $k$ character-perturbed copies of the input. \textbf{Visual TTA} applies $k$ image transformations (rotation $\pm 3^\circ$, contrast and brightness $\pm 5\%$) on multi-modal inputs.

We evaluate each method with $k \in \{2, 4, 6\}$, and extend $k$ to 10 for ablations. Because the best $k$ varies by task, we select it per method and dataset on a held-out sample drawn separately from the evaluation subset. On multi-modal datasets we also combine visual and semantic TTA.

\subsection{Evaluation}

We employ task-appropriate accuracy metrics for each benchmark. For MMLU, MMMLU, MMMU and HLE, we measure the fraction of correctly answered questions, parsing each response to extract the answer and comparing it with the ground truth. For IMDB, we compute binary classification accuracy of the predicted sentiment. Math500 needs additional verification to assess semantic equivalence: we apply string matching on normalized LaTeX expressions, followed by symbolic equivalence checking with SymPy \citep{sympy} to catch equivalent answers written differently.

All methods incur cost proportional to $k$. Semantic TTA pays extra for generating rephrasings, while lexical augmentation is nearly free, and Section~\ref{sec_cost_accuracy} analyzes the resulting trade-offs. All experiments run on Amazon Bedrock with a fixed seed, priced on demand as of June 2026.

\section{Results and Analysis}
\label{sec_results}

\subsection{Main Results}
\label{sec_main_results}

\begin{figure*}[!t]
\centering
\includegraphics[width=\textwidth,trim={0 0 0 0},clip]{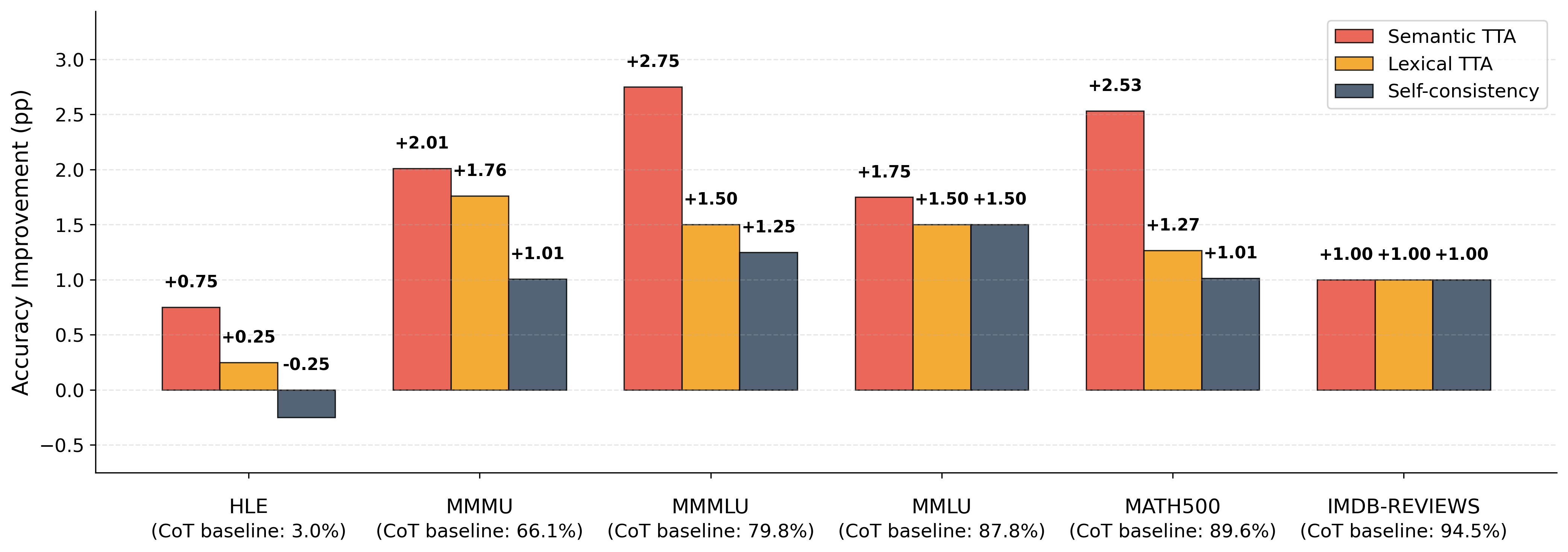}
\vspace{-1ex}
\caption{Accuracy improvement over the CoT baseline (in pp). Semantic TTA achieves the highest gains across most benchmarks, outperforming the self-consistency baseline on five of six tasks. The number of samples for each method is selected via grid search over $k \in \{2, 4, 6\}$.}
\label{fig:summary}
\vspace{-2ex}
\end{figure*}

Figure~\ref{fig:summary} reports the accuracy improvement of each TTA method and the self-consistency baseline over single-call CoT prompting for each dataset. Across all six benchmarks, semantic TTA delivers the largest average gain of 1.8 percentage points (pp), and outperforms self-consistency on five of six benchmarks. The gap over self-consistency is largest on Math500 and MMMLU (+1.52 and +1.50 pp), where aggregating across alternative phrasings overcomes prompt-formulation sensitivity that repeated sampling does not address. Lexical TTA gains less (1.2 pp on average), suggesting that character-level perturbations add noise that partially offsets the aggregation benefit.

Self-consistency improves over single-call CoT but is consistently outperformed by semantic TTA. This indicates that gains from aggregation have two sources: variance reduction from repeated sampling (captured by self-consistency), and additional diversity introduced by varying the input (captured by input-side TTA). Semantic TTA benefits from both, and Appendix~\ref{app:variance} confirms that every aggregation method reduces prediction variance relative to a single call.

How much diversity each method generates does not track what it gains. Measuring how often the aggregated candidates fail to agree, semantic TTA's disagree on 27.4\% of questions at matched $k$ against 22.8\% for self-consistency, so rewording does reach candidates that resampling one wording does not. Lexical TTA disagrees most of all at 37.8\% while gaining less accuracy, and perturbing two modalities at once (Section~\ref{sec_modality}) falls below the single-call baseline. More disagreement is therefore not better disagreement. What separates the methods is whether the perturbation keeps the meaning intact: paraphrases change which reading stands out while leaving the answer unchanged, so each candidate is plausible and voting can choose between them, whereas character corruption and stacked multi-modal noise produce variation that is often simply wrong. Appendix~\ref{app:qualitative} reports the disagreement rates per dataset alongside a worked example where the original phrasing of a question yields the wrong answer while more explicit phrasings of it recover the right one.

\begin{figure*}[b]
\centering
\includegraphics[width=\textwidth,trim={0 0 2.5cm 0},clip]{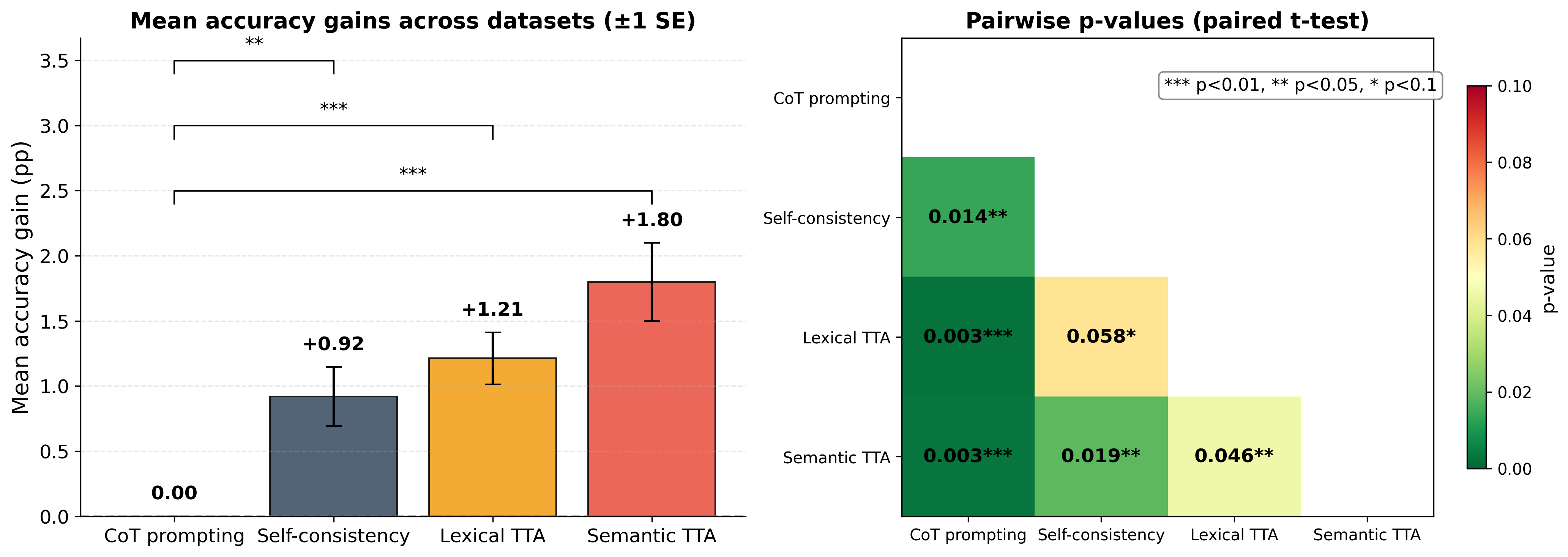}
\caption{Pairwise statistical significance of accuracy gains across datasets. Semantic TTA shows statistically significant improvements over CoT and over self-consistency ($p < 0.05$).}
\label{fig:significance}
\end{figure*}

Figure~\ref{fig:significance} reports paired $t$-tests on mean accuracy gains aggregated across all six datasets. Semantic TTA achieves statistically significant improvements over both single-call CoT ($p < 0.01$) and self-consistency ($p < 0.05$). Lexical TTA also significantly outperforms single-call CoT ($p < 0.01$), as does self-consistency ($p < 0.05$), but their gains are not significantly different from each other at the 5\% level. A non-parametric paired bootstrap over the pooled per-question gains ($n=2{,}400$) confirms these conclusions, with the 95\% confidence interval for semantic TTA well above zero at $[+0.88, +2.71]$ pp (see Appendix~\ref{app:bootstrap} for details).

\subsection{Cost-Accuracy Trade-offs}
\label{sec_cost_accuracy}

\begin{figure}[t]\centering
\begin{subfigure}[b]{0.45\columnwidth}
  \centering
  \includegraphics[width=\columnwidth,trim={0 0 0 0},clip]{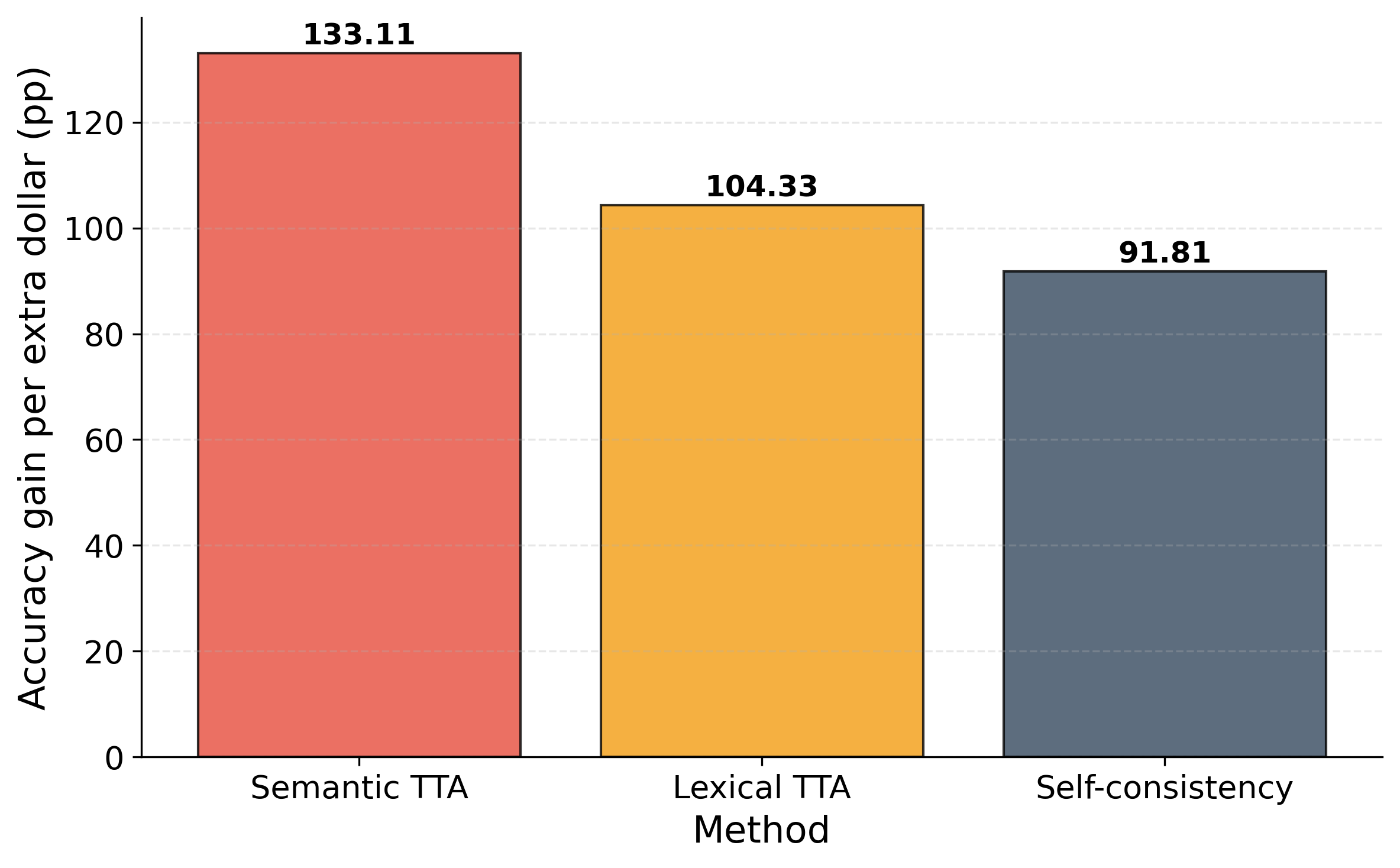}
  \caption{Accuracy gain per extra dollar spent.}
\end{subfigure}
\hfill
\begin{subfigure}[b]{0.45\columnwidth}
  \centering
  \includegraphics[width=\columnwidth,trim={0 0 0 0},clip]{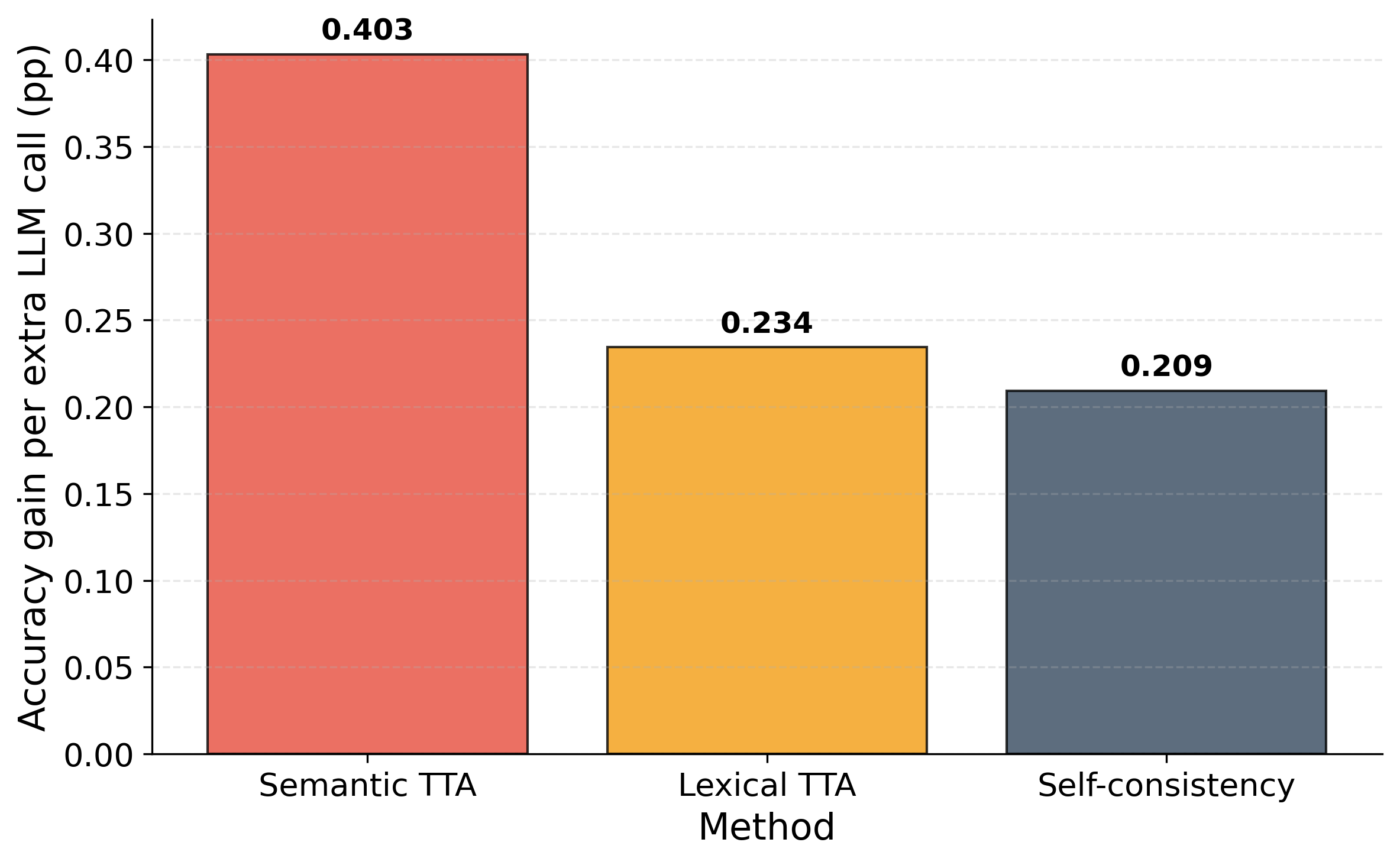}
  \caption{Accuracy gain per extra LLM call.}
\end{subfigure}
\caption{Cost-effectiveness of each method, measured as accuracy gain relative to additional cost (left) and number of LLM calls (right). Semantic TTA is most effective on both.}
\label{fig:effectiveness}
\end{figure}

Varying the input costs slightly more than resampling it, since semantic TTA spends one extra call on rephrasing. Here, we ask whether the input-side gains justify the additional cost. Figure~\ref{fig:effectiveness} compares accuracy gain per unit cost and per LLM call across methods. Semantic TTA is the most cost-effective option on both measures, delivering roughly 1.5$\times$ more accuracy per dollar and 1.9$\times$ more per additional LLM call than self-consistency. Per-dataset cost-accuracy frontiers, reported in Appendix~\ref{app:frontiers}, show the same ordering: semantic TTA achieves the highest accuracy at nearly every cost level, while lexical TTA traces a cheaper frontier close to self-consistency, making it a reasonable fallback when even small augmentation overhead is undesirable.

While statistically significant, we note that gains of 1--2pp may not justify a 2--6$\times$ cost increase in all settings. Therefore, TTA is most valuable when baseline accuracy is moderate (40--80\%).

\subsection{Number of Samples}

\begin{figure*}[!t]
\centering
\begin{minipage}[b]{0.44\textwidth}
  \centering
  \footnotesize
  \renewcommand{\arraystretch}{1.1}
  \begin{tabular*}{\linewidth}{@{\extracolsep{\fill}}lccc@{}}
  \toprule
  \multirow{2}{*}{Dataset} & Semantic & Lexical & Self- \\
          & TTA      & TTA     & consistency \\
  \midrule
  HLE     & 6 & 6 & \textbf{2} \\
  IMDB    & 4 & \textbf{2} & 6 \\
  Math500 & \textbf{4} & 6 & 6 \\
  MMLU    & \textbf{4} & 6 & 6 \\
  MMMLU   & 6 & 6 & \textbf{2} \\
  MMMU    & \textbf{2} & 6 & 6 \\
  \midrule
  Average & \textbf{4.33} & 5.33 & 4.67 \\
  \bottomrule
  \end{tabular*}
  \captionof{table}{Optimal number of samples $k$ per dataset and method, selected on a held-out sample via grid search over $k \in \{2, 4, 6\}$. Bold marks the smallest optimal $k$ in each row. Semantic TTA peaks at smaller $k$ on average, reaching its best accuracy with less compute than other methods.}
  \label{tab:optimal_k}
\end{minipage}
\hfill
\begin{minipage}[b]{0.53\textwidth}
  \centering
  \includegraphics[width=\linewidth,trim={0 -0.22cm 0 0},clip]{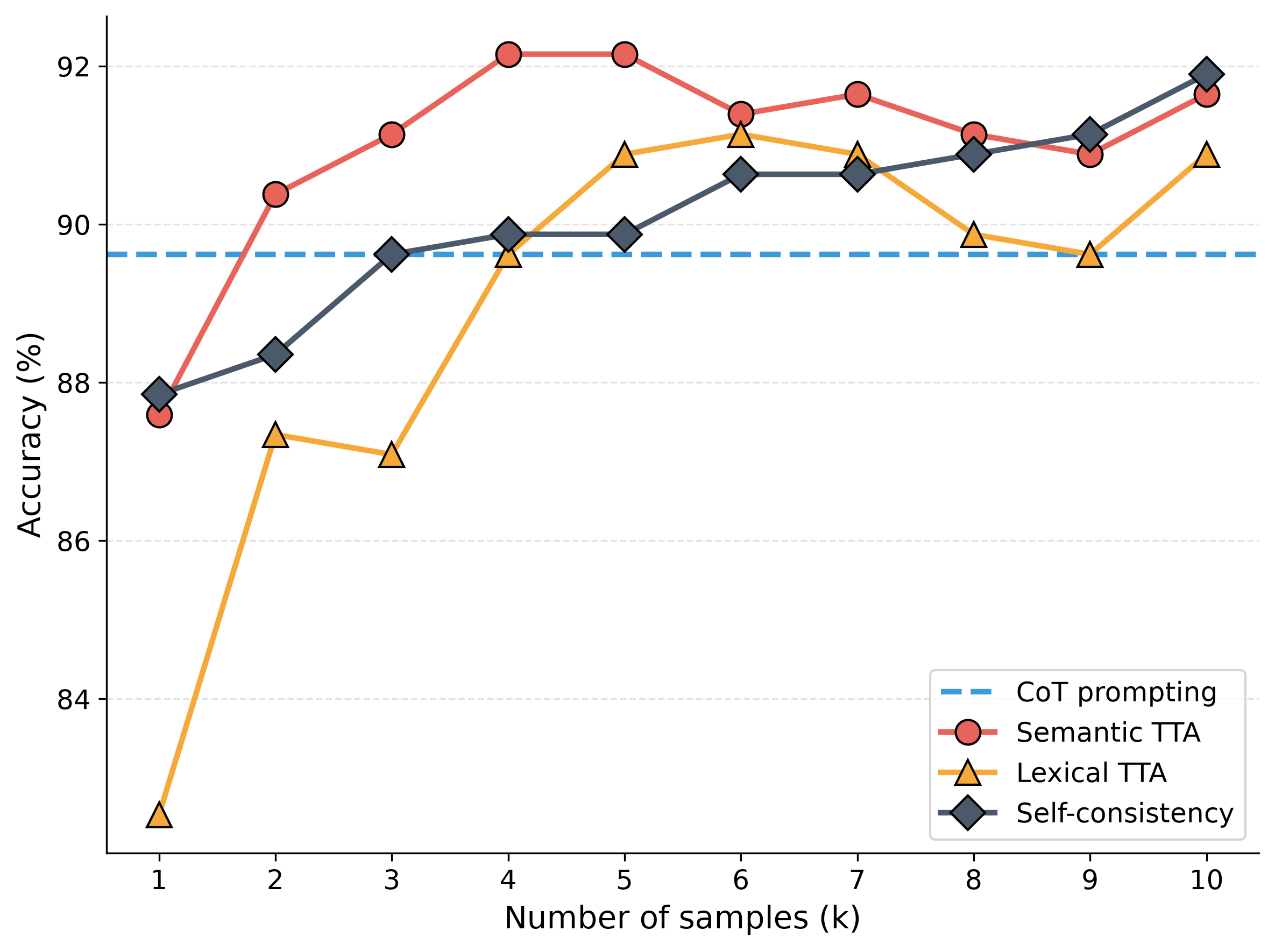}
  \captionof{figure}{Effect of the number of samples on Math500. Semantic TTA peaks at $k=5$; self-consistency continues to improve up to $k=10$.}
  \label{fig:num_augmentations}
\end{minipage}
\end{figure*}

Table~\ref{tab:optimal_k} reports the optimal $k$ per dataset and method, where $k$ counts the responses aggregated: augmented inputs for TTA, repeated samples for self-consistency. Semantic TTA peaks with fewer samples on average ($k=4.33$), indicating higher-quality diversity than the alternatives.

Figure~\ref{fig:num_augmentations} shows extended experiments on Math500 with $k$ up to 10. Accuracy is non-monotonic in $k$: small drops at certain $k$ reflect stochastic paraphrase generation and tie-breaking dynamics, particularly for odd vs. even $k$. These fluctuations are small (within 1 pp). Semantic TTA peaks at $k=5$ with diminishing returns thereafter; self-consistency continues to improve up to $k=10$, consistent with \citet{wang2022selfconsistency}'s finding that self-consistency benefits from larger sample counts, and at $k=10$ it nearly matches semantic TTA on Math500. Based on these results, we recommend starting with $k=4$ for semantic TTA, balancing accuracy and cost.

\subsection{Augmentation Modality}
\label{sec_modality}

For multi-modal tasks, we examine whether text or image augmentation is more effective. Each MMMU example includes a textual question and at least one image. This allows us to compare augmenting text, images, or both. Table~\ref{tab:mmmu_multi-modal} reports results on MMMU.

Text-based semantic TTA achieves the highest accuracy (68.09\%) with $k=2$, outperforming visual TTA (67.59\%) at $k=6$. This suggests that LLM predictions are more sensitive to question phrasing than to the mild visual perturbations we apply: the model's image understanding is already robust to the small rotations and brightness or contrast shifts we test. We emphasize that this conclusion is scoped to these mild photometric and geometric transforms; stronger or structurally different visual augmentations may behave differently and are left to future work. Visual TTA still beats self-consistency (67.09\%), so input-side diversity helps even when applied to the image alone.

We combine the two modalities in two ways: the \emph{combined} variant perturbs both the question and the image in every variant, while the \emph{alternating} variant perturbs only one modality per variant, chosen at random. Neither helps. The combined variant (65.08\%) falls below single-call CoT (66.08\%), indicating that perturbing both modalities at once accumulates enough noise to outweigh the benefits of aggregation, and alternating (66.33\%) barely improves on the baseline because each variant dilutes the more useful text perturbation with image-only ones. For multi-modal tasks, we therefore recommend text-based semantic augmentation alone.

\begin{table}[b]
  \centering
  \small
  \caption{Multi-modal results on MMMU. Visual TTA improves over baseline but falls behind text-based TTA. The $k$ is selected on a held-out sample via grid search over $k \in \{2, 4, 6\}$.}
  \label{tab:mmmu_multi-modal}
  \begin{tabular*}{\columnwidth}{@{\extracolsep{\fill}}llcc@{}}
    \toprule
    Modality & Method & Selected $k$ & Accuracy (\%) \\
    \midrule
    \multirow{2}{*}{--} & CoT prompting & -- & 66.08 \\
                        & Self-consistency & 6 & 67.09 \\
    \midrule
    \multirow{2}{*}{Text} & Lexical TTA           & 6 & 67.84 \\
                          & \textbf{Semantic TTA} & \textbf{2} & \textbf{68.09} \\
    \midrule
    Image & Visual TTA & 6 & 67.59 \\
    \midrule
    \multirow{2}{*}{Text + Image} & Alternating & 2 & 66.33 \\
                                  & Combined   & 4 & 65.08 \\
    \bottomrule
  \end{tabular*}
\end{table}

\subsection{Base Model Choice}
\label{sec_base_model_size}

We examine whether TTA benefits vary across model sizes by evaluating all methods on Claude 4.5 Haiku, Sonnet, and Opus on MMMLU. Table~\ref{tab:cross_model} reports the results.

Semantic TTA achieves the highest accuracy at every model size. The magnitude of improvement decreases with model scale: Haiku gains 2.75pp, Sonnet -- 2pp, and Opus -- only 0.25pp. This pattern suggests that TTA provides larger benefits when baseline accuracy leaves more room for improvement. For high-performing LLMs already near ceiling, TTA offers marginal gains as the model is already robust to prompt variations. The same pattern is even more pronounced for lexical TTA, which improves Haiku and Sonnet but degrades Opus (87.00\% vs.~88.25\% baseline), indicating that character-level noise can hurt strong models whose predictions are already close to ceiling.

Comparing across tiers, Haiku with semantic TTA (82.50\%) does not match Sonnet's single-call accuracy (85.50\%), and Sonnet with TTA (87.50\%) remains below Opus's single-call accuracy (88.25\%). TTA is therefore not a substitute for upgrading to a stronger model when one is available within budget. Rather than a limitation, this shows where TTA belongs: it is a tool for smaller models, most valuable when a larger one is unavailable or prohibitively expensive and more accuracy must come from a fixed base model.

Because the rephraser and the answering model may be different, we also test whether delegating augmentation to a stronger model helps. It does not reliably, as Haiku rephrasing its own inputs matches or beats Sonnet and Opus rephrasers on three of six benchmarks at a third to a fifth of the augmentation cost (Appendix~\ref{app:rephraser}), so the model already serving the task can augment its own inputs.

\begin{table*}[!t]
  \centering
  \small
  \caption{Performance across the base model sizes on MMMLU. Semantic TTA shows the largest gains on smaller models. Gains diminish as base accuracy approaches ceiling.}
  \label{tab:cross_model}
  \begin{tabular*}{\columnwidth}{@{\extracolsep{\fill}}lccc@{}}
  \toprule
  Method & Claude 4.5 Haiku & Claude 4.5 Sonnet & Claude 4.5 Opus \\
  \midrule
  CoT prompting & 79.75 & 85.50 & 88.25 \\
  Self-consistency & 81.00 & 86.75 & 88.25 \\
  Lexical TTA  & 81.25 & 86.75 & 87.00 \\
  \textbf{Semantic TTA} & \textbf{82.50} & \textbf{87.50} & \textbf{88.50} \\
  \midrule
  Semantic TTA gain over CoT & +2.75 & +2.00 & +0.25 \\
  \bottomrule
  \end{tabular*}
\end{table*}

%
%

\section{Conclusion}
\label{sec_conclusion}

This paper asks where diversity should be injected when several LLM responses are aggregated at inference time. We study Test-Time Augmentation (TTA), which varies the input and aggregates the resulting predictions, evaluating semantic rephrasing, lexical perturbations, and visual transformations across six benchmarks against CoT prompting and self-consistency.

Several key findings emerge from our experiments. Where the diversity comes from matters more than how much of it there is. Rewording the question gives the largest gains, beating self-consistency on five of six benchmarks at a lower cost per point of accuracy, and both semantic and lexical TTA improve significantly on single-call CoT prompting. Yet the ranking does not follow the amount of variation produced: character-level noise makes the aggregated candidates disagree more often than paraphrasing does while gaining less, and perturbing the question and the image together falls below the single-call baseline. What matters is whether the perturbation keeps the meaning intact, because only then is the disagreement it creates something that majority voting can use.

Our ablations yield practical guidance. Semantic TTA reaches near-optimal accuracy at $k=4$, while self-consistency keeps improving at larger $k$. Gains are larger for smaller LLMs whose baseline leaves more room, and lexical TTA can even degrade strong models already near ceiling. Delegating rephrasing to a stronger LLM does not help, so the LLM serving the task can augment its own inputs.

TTA proves to be a simple way to extract more accuracy from a fixed base model, most useful when a stronger LLM is unavailable or too expensive. More broadly, the linguistic form of an input is not just a weakness to minimize but a design axis that can be varied at inference time. We hope these findings encourage further work treating surface form as something to use rather than only to control.

Our study has several limitations. First, TTA applies only to tasks with a well-defined notion of answer equivalence, where majority voting is meaningful; open-ended generation such as summarization would require different aggregation mechanisms. Second, our evidence covers a single proprietary model family: patterns are consistent across Claude Haiku, Sonnet, and Opus, but behavior on open-weight families remains to be verified, so claims about LLMs in general should be read as claims about current small and mid-tier models. Third, our multilingual evidence aggregates over the fourteen languages of MMMLU, and paraphrase quality varies by language, so gains may not transfer uniformly to lower-resource languages. Fourth, majority voting can amplify confidently wrong answers, so aggregated predictions should not be treated as calibrated confidence estimates in high-stakes settings. Finally, since TTA extends rather than replaces self-consistency, tuning the balance of input-side and output-side diversity is a natural direction for future work.

%
%

\bibliography{references}
\bibliographystyle{plainnat}

%
%

\clearpage
\appendix
%
%

\clearpage
\appendix
\section{Prompt Templates}
\label{app:prompts}

\subsection{Answering and Rephrasing Prompts}
\label{app:prompts:tta}

Section~\ref{app:prompts:tta} presents the answering and rephrasing prompts. The answering prompt is used by both the baseline and all TTA methods. The rephrasing prompt is used exclusively by semantic TTA to generate question variations. 

\begin{tcolorbox}[title=\textbf{Answering}, enhanced, breakable, colback=gray!5, colframe=gray!50!black, fonttitle=\bfseries]
Answer the following question. First, think through the problem step-by-step, then provide your final answer.

Guidelines:
\begin{itemize}
    \item Think through your reasoning in <thinking></thinking> tags (use maximum 200 words)
    \item Provide your final answer inside <answer></answer> XML tags
    \begin{itemize}
        \item For multiple-choice questions: Output only the letter (A, B, C, or D) in the answer tags
        \item For mathematical questions: Output only the numerical answer or mathematical expression in the answer tags (e.g., <answer>16</answer> or <answer>\textbackslash frac\{1\}\{2\}</answer>)
        \item For open-ended questions: Output the specific answer requested in the answer tags
    \end{itemize}
\end{itemize}

Question:\\
<question>\\
\{question\}\\
</question>

Your response format:\\
<thinking>\\
Your step-by-step reasoning in less than 200 words\\
</thinking>

<answer>\\
Your final answer here\\
</answer>
\end{tcolorbox}

\begin{tcolorbox}[title=\textbf{Rephrasing}, enhanced, breakable, colback=gray!5, colframe=gray!50!black, fonttitle=\bfseries]
You are given a question that may be either:
\begin{itemize}
    \item A multiple-choice question with answer options (A, B, C, D)
    \item An open-ended question requiring a specific answer
\end{itemize}

Your task is to rephrase this question into \{num\_augmentations\} different variations. Each variation should:
\begin{itemize}
    \item Preserve the exact same meaning and correct answer
    \item Use different wording or sentence structure
    \item If there are answer choices, maintain them exactly as they are (same letters, same options)
    \item If the question includes specific output format instructions (e.g., ``output as array'', ``use XML tags''), preserve these EXACTLY as written - do not paraphrase formatting requirements
    \item If the question references images (e.g., ``<image 1>'', ``in the diagram''), maintain these references in the same form
    \item Keep all the essential information needed to answer the question
\end{itemize}

Output each rephrased question in XML tags: <q1></q1>, <q2></q2>, <q3></q3>, etc.

Do not explain your thinking. Do not add any information to your answer except for the rephrased questions.

Original Question:\\
<question>\\
\{question\}\\
</question>

Generate \{num\_augmentations\} rephrased variations now:
\end{tcolorbox}

\subsection{Dataset-Specific Prompts}
\label{app:prompts:dataset}

Section~\ref{app:prompts:dataset} presents dataset-specific prompts that are inserted into the \texttt{\{question\}} placeholder of the answering and rephrasing templates.

\begin{tcolorbox}[title=\textbf{Math500}, enhanced, breakable, colback=gray!5, colframe=gray!50!black, fonttitle=\bfseries]
\{problem\}

Provide your final answer in the simplest form.
\end{tcolorbox}

\begin{tcolorbox}[title=\textbf{IMDB Reviews}, enhanced, breakable, colback=gray!5, colframe=gray!50!black, fonttitle=\bfseries]
Classify the sentiment of the following review as either `positive' or `negative'.

Review: \{review\}
\end{tcolorbox}

\begin{tcolorbox}[title=\textbf{MMLU / MMMLU}, enhanced, breakable, colback=gray!5, colframe=gray!50!black, fonttitle=\bfseries]
\{question\}

A. \{option\_a\}\\
B. \{option\_b\}\\
C. \{option\_c\}\\
D. \{option\_d\}
\end{tcolorbox}

\begin{tcolorbox}[title=\textbf{MMMU}, enhanced, breakable, colback=gray!5, colframe=gray!50!black, fonttitle=\bfseries]
\{question\}

A. \{option\_a\}\\
B. \{option\_b\}\\
C. \{option\_c\}\\
D. \{option\_d\}

[Images are provided as visual input to the model]
\end{tcolorbox}

\begin{tcolorbox}[title=\textbf{HLE (Multiple Choice)}, enhanced, breakable, colback=gray!5, colframe=gray!50!black, fonttitle=\bfseries]
\{question\}

Provide your answer as a single letter (A, B, C, or D).

[Images are provided as visual input to the model when available]
\end{tcolorbox}

\begin{tcolorbox}[title=\textbf{HLE (Exact Match)}, enhanced, breakable, colback=gray!5, colframe=gray!50!black, fonttitle=\bfseries]
\{question\}

Provide your answer in the simplest form.

[Images are provided as visual input to the model when available]
\end{tcolorbox}

\section{Variance and Significance Analysis}
\label{app:variance}

Because the accuracy gains we report are on the order of 1 to 2 pp, this section examines them from two complementary angles: whether aggregation reduces prediction variance as the method intends, and whether the resulting gains are statistically distinguishable from zero under a non-parametric test.

\subsection{Prediction Variance}

Table~\ref{tab:variance} reports the standard deviation of prediction-level accuracy across dataset questions for each method. All aggregation methods reduce variance compared to single-call CoT prompting, with semantic TTA achieving the lowest variance on most benchmarks. The exception is HLE, where semantic TTA increases variance (17.06\% $\rightarrow$ 19.00\%). This likely stems from HLE's very low baseline accuracy (3\%), where paraphrasing introduces additional variation without helping the model answer questions it fundamentally cannot solve.

\begin{table}[!t]
  \centering
  \small
  \caption{Standard deviation (\%) of predictions across datasets and methods.}
  \label{tab:variance}
  \begin{tabular*}{\textwidth}{@{\extracolsep{\fill}}lcccc@{}}
  \toprule
  Dataset & CoT prompting & Semantic TTA & Lexical TTA & Self-consistency \\
  \midrule
  HLE          & 17.06 & 19.00 & 17.73 & \textbf{16.35} \\
  IMDB         & 22.80 & \textbf{20.73} & \textbf{20.73} & \textbf{20.73} \\
  Math500      & 30.50 & \textbf{26.89} & 28.78 & 27.29 \\
  MMLU         & 32.79 & \textbf{30.66} & \textbf{30.66} & \textbf{30.66} \\
  MMMLU        & 40.19 & \textbf{38.00} & 39.03 & 39.23 \\
  MMMU         & 47.34 & \textbf{46.61} & 46.71 & 46.99 \\
  \bottomrule
  \end{tabular*}
\end{table}

\subsection{Bootstrap Significance}
\label{app:bootstrap}

As a non-parametric complement to the paired $t$-tests in Section~\ref{sec_main_results}, we run a paired bootstrap over the per-question accuracy gains relative to single-call CoT prompting. We pool the per-question gains across all six datasets ($n=2{,}400$) and resample with replacement (5{,}000 resamples) to obtain a 95\% confidence interval on the mean gain for each method. Table~\ref{tab:bootstrap} reports the results.

The bootstrap confirms the parametric conclusions. Semantic TTA has the largest mean gain (+1.79 pp) and its confidence interval sits well above zero, matching the +1.8 pp average reported in Section~\ref{sec_main_results}. Both lexical TTA and self-consistency also clear zero, with the consistent ordering: input-side semantic TTA sits furthest above the noise floor, while self-consistency is the weakest, with its lower confidence bound only marginally positive.

\begin{table}[!t]
  \centering
  \small
  \caption{Paired bootstrap on pooled per-question accuracy gains over single-call CoT prompting ($n=2{,}400$, 5{,}000 resamples). All methods clear zero; semantic TTA has the highest $t$ value.}
  \label{tab:bootstrap}
  \begin{tabular*}{\textwidth}{@{\extracolsep{\fill}}lccc@{}}
  \toprule
  Method & Mean gain (pp) & 95\% CI (pp) & $t$ \\
  \midrule
  \textbf{Semantic TTA} & \textbf{+1.79} & $[+0.88, +2.71]$ & \textbf{3.86} \\
  Lexical TTA & +1.25 & $[+0.25, +2.21]$ & 2.47 \\
  Self-consistency & +0.92 & $[+0.04, +1.79]$ & 2.06 \\
  \bottomrule
  \end{tabular*}
\end{table}

\section{Qualitative Analysis}
\label{app:qualitative}

To illustrate how input-side diversity changes model behavior, we inspect individual predictions and quantify how often the aggregated candidates disagree with one another.

\subsection{Worked Example}

The example below is an MMLU economics question on which single-call CoT prompting and self-consistency both answer incorrectly, while semantic TTA recovers the correct answer. The original phrasing is a bare sentence fragment (``Tariffs and quotas'') that leaves the intended comparison implicit, and the model settles on option D. Two of the four rephrasings state the question explicitly in terms of the quantity sold domestically, and under those phrasings the model answers C. Aggregation resolves the split in favor of the correct answer.

\begin{tcolorbox}[title=\textbf{MMLU question where semantic TTA corrects the prediction}, enhanced, breakable, colback=gray!5, colframe=gray!50!black, fonttitle=\bfseries]
\textbf{Original question} (answer: C)\\
Tariffs and quotas\\
A. result in lower domestic prices.\\
B. sometimes raise and sometimes lower the amount of the product sold domestically.\\
C. lower the amount of the product sold domestically.\\
D. raise the amount of the product sold domestically.

\medskip
\textbf{Predictions per input variant} (answer choices identical throughout, omitted for brevity)

\smallskip
\begin{tabular}{@{}p{0.80\textwidth}cc@{}}
\emph{Original:} Tariffs and quotas & D & \xmark \\
\emph{Variant 1:} Tariffs and quotas: & D & \xmark \\
\emph{Variant 2:} Which statement accurately describes the effects of tariffs and quotas? & D & \xmark \\
\emph{Variant 3:} What is the typical consequence of implementing tariffs and quotas on domestic sales? & C & \cmark \\
\emph{Variant 4:} How do tariffs and quotas generally affect the quantity of products sold within the domestic market? & C & \cmark \\
\end{tabular}

\medskip
\textbf{Aggregation.} Votes $\{\text{C}: 3, \text{D}: 2\}$ $\rightarrow$ final answer \textbf{C} \cmark\\
For reference, single-call CoT prompting answers D, and self-consistency answers D at both $k=4$ and $k=6$.
\end{tcolorbox}

The split is linguistically systematic rather than arbitrary. The two variants that recover the correct answer are the two that make the dimension of comparison explicit (``domestic sales'', ``quantity of products sold''), supplying the argument that the elliptical original leaves unsaid, whereas the two that merely convert the fragment into a well-formed question inherit its ambiguity and repeat the original error. No variant adds information beyond what the original question and its options contain; what changes is which reading the wording brings out. This is the mechanism the paper posits: surface form redistributes probability mass across answers, and aggregation exploits it.

\subsection{Candidate Disagreement Rates}
\label{app:disagreement}

The worked example raises the question of whether input perturbation produces measurably more candidate diversity than resampling the same input. We quantify this by recording, for every question, whether the $k+1$ aggregated candidates were unanimous or split. Table~\ref{tab:disagreement} reports the resulting disagreement rate at $k=4$ on the five benchmarks with categorical answers; IMDB is omitted because its near-ceiling binary accuracy leaves almost no disagreement to measure.

Semantic TTA's candidates disagree on 27.4\% of questions against 22.8\% for self-consistency, and the ordering is consistent across all five datasets. Input perturbation therefore generates genuinely more diverse candidates than sampling repeated reasoning paths from a fixed input, which is the premise of the accuracy gains reported in Section~\ref{sec_main_results}. Lexical TTA produces the highest disagreement (37.8\%) yet not the highest accuracy, confirming that diversity alone is insufficient: character-level noise generates variation that is less often correct, which is why it trails semantic rephrasing despite perturbing the input more aggressively.

\begin{table}[!t]
  \centering
  \small
  \caption{Candidate disagreement rate (\%) at $k=4$: the fraction of questions on which the aggregated candidates are not unanimous, computed on a matched question set per dataset. Higher disagreement means more candidate diversity, not better accuracy.}
  \label{tab:disagreement}
  \begin{tabular*}{\textwidth}{@{\extracolsep{\fill}}lccc@{}}
  \toprule
  Dataset & Semantic TTA & Lexical TTA & Self-consistency \\
  \midrule
  HLE     & 70.0 & 70.9 & 61.6 \\
  Math500 &  8.3 & 38.0 &  7.4 \\
  MMLU    & 13.5 & 19.8 & 10.1 \\
  MMMLU   & 24.6 & 38.4 & 19.7 \\
  MMMU    & 36.6 & 48.3 & 31.9 \\
  \midrule
  Pooled  & 27.4 & 37.8 & 22.8 \\
  \bottomrule
  \end{tabular*}
\end{table}

\section{Decoupled Rephraser Analysis}
\label{app:rephraser}

Semantic TTA uses the answering model to rephrase its own inputs, but the two roles need not be filled by the same model. Because a stronger rephraser might produce more useful input-side diversity, we test this directly: Table~\ref{tab:rephraser} compares Haiku rephrasing its own inputs against Sonnet and Opus acting as rephrasers while Haiku still answers, across all six benchmarks.

A stronger rephraser does not reliably help. Self-rephrasing is at least as accurate as both alternatives on three of six benchmarks, and the three averages fall within 0.4 pp of each other. Where a stronger rephraser wins, the margin is small (at most +0.8 pp, on MMMLU) and comparable to the margin by which self-rephrasing wins elsewhere (+0.8 pp on MMMU), while costing three to five times more per augmentation call. Paraphrasing is evidently easy enough that mid-tier capability suffices, which is convenient for deployment: TTA needs only the model already serving the task, with no second endpoint to provision. It also shows the gains are not an artifact of distilling a stronger model's understanding through the rephrasings, since they hold when the model augments its own inputs.

\begin{table}[!t]
  \centering
  \small
  \caption{Effect of decoupling the rephraser from the answering model. All columns answer with Claude 4.5 Haiku and differ only in which model generates the rephrasings; accuracy (\%) is the best over $k \in \{2, 4, 6\}$. A stronger rephraser does not reliably improve accuracy.}
  \label{tab:rephraser}
  \begin{tabular*}{\textwidth}{@{\extracolsep{\fill}}lccc@{}}
  \toprule
  Dataset & Haiku (self-rephrase) & Sonnet rephraser & Opus rephraser \\
  \midrule
  HLE     & \textbf{4.10}  & 3.75  & 3.25 \\
  IMDB    & 95.10 & \textbf{95.50} & \textbf{95.50} \\
  Math500 & 91.40 & \textbf{91.50} & 91.25 \\
  MMLU    & \textbf{89.20} & 88.75 & 88.50 \\
  MMMLU   & 81.70 & \textbf{82.50} & 82.00 \\
  MMMU    & \textbf{67.56} & 66.75 & 66.25 \\
  \midrule
  Average & \textbf{71.51} & 71.46 & 71.12 \\
  \bottomrule
  \end{tabular*}
\end{table}

\section{Cost-Accuracy Frontiers}
\label{app:frontiers}

\begin{figure}[!ht]\centering
\begin{subfigure}[b]{0.42\columnwidth}
  \centering
  \includegraphics[width=\columnwidth,trim={0 0 0 0},clip]{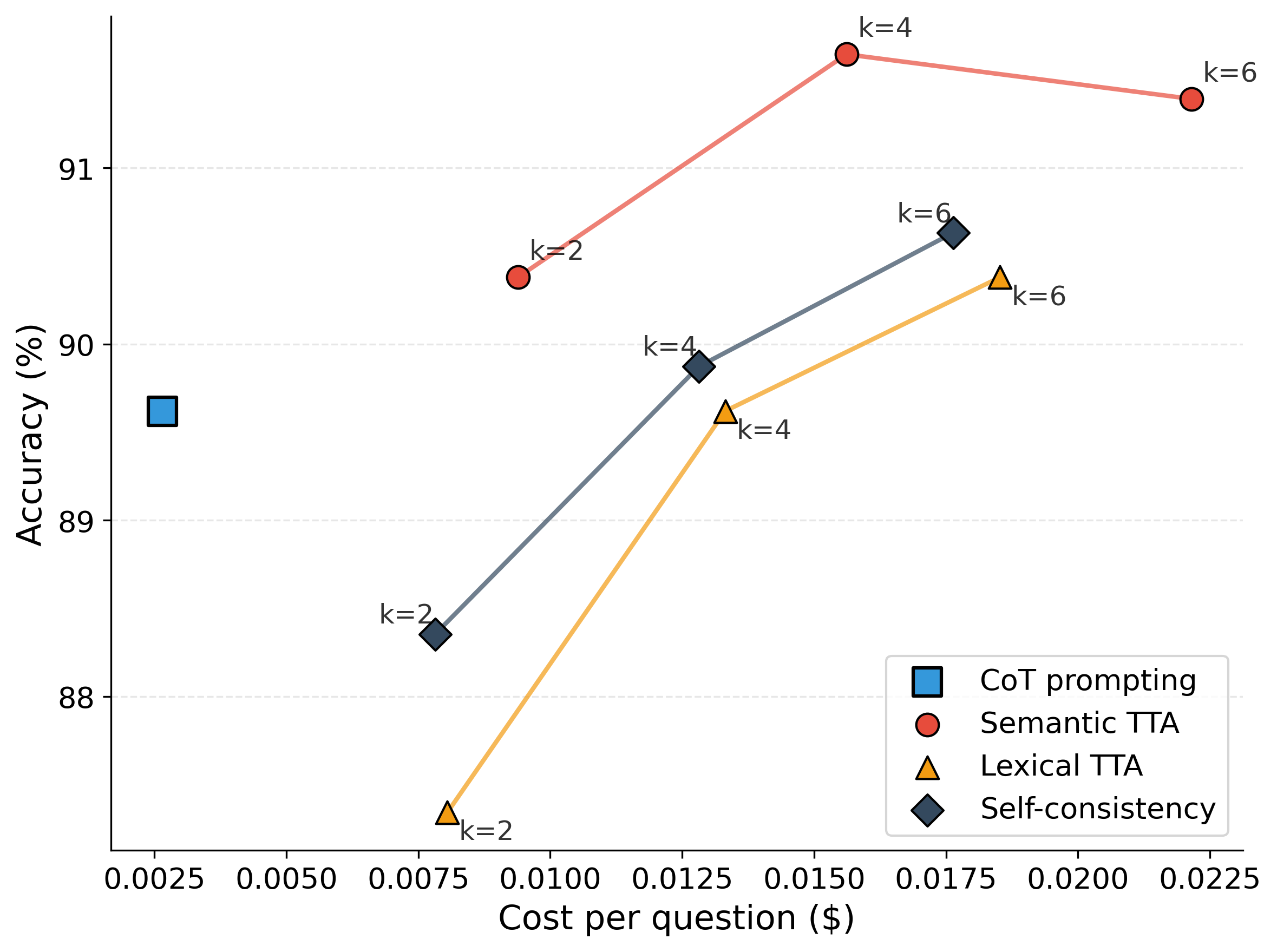}
  \caption{Math500}
\end{subfigure}
\hfill
\begin{subfigure}[b]{0.42\columnwidth}
  \centering
  \includegraphics[width=\columnwidth,trim={0 0 0 0},clip]{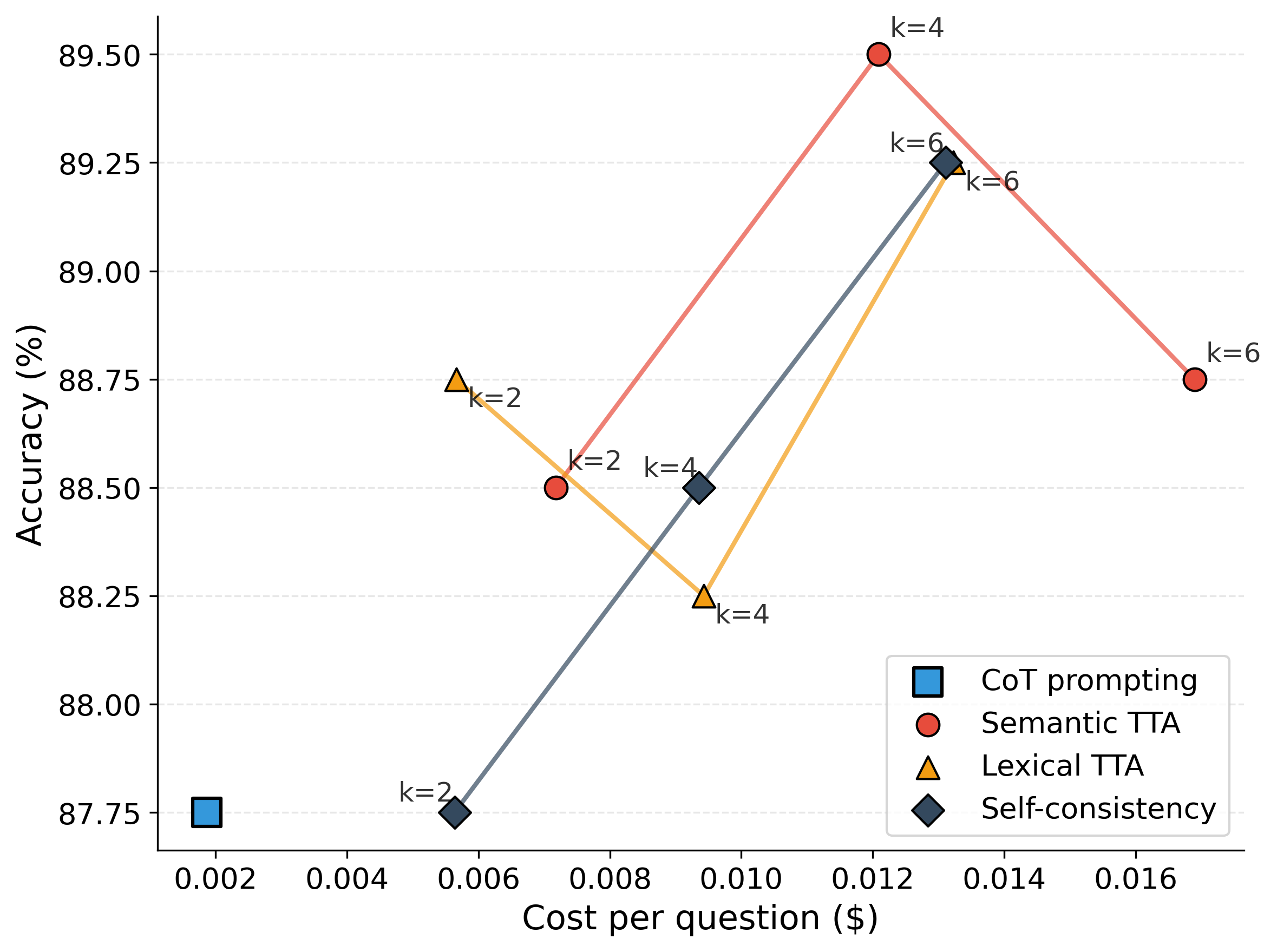}
  \caption{MMLU}
\end{subfigure}
\caption{Cost-accuracy frontiers on Math500 and MMLU. Points are labelled with the number of augmentations $k$; cost per question includes the rephrasing call for semantic TTA.}
\label{fig:frontiers}
\end{figure}

Figure~\ref{fig:frontiers} plots accuracy against cost per question on Math500 and MMLU, tracing each method's frontier as $k$ grows. Semantic TTA sits above the other methods at nearly every cost level on both datasets, the exception being a drop at $k=6$ on MMLU, and it reaches peak accuracy at $k=4$ where self-consistency needs more samples to come close. Lexical TTA traces a frontier close to self-consistency at a lower cost, since it requires no rephrasing call. The cost axis already includes semantic TTA's rephrasing call, so its advantage is net of that overhead.

%
%


\end{document}